\documentclass[conference]{IEEEtran}
\usepackage{times}
\usepackage[dvipsnames]{xcolor}

\usepackage[numbers]{natbib}
\usepackage{multicol}
\usepackage{hyperref}
\usepackage[dvipsnames]{xcolor}
\hypersetup{
    colorlinks=true,
    linkcolor=blue,
    filecolor=magenta,      
    urlcolor=StanfordRed
    }

\usepackage{graphics} 
\usepackage{epsfig} 
\usepackage{mathptmx} 
\usepackage{times} 
\usepackage{amsmath} 
\usepackage{amssymb}  
\usepackage{siunitx} 
\usepackage{textcomp}
\usepackage{gensymb} 
\usepackage{booktabs}
\usepackage{multirow}
\usepackage{xcolor}
\usepackage{cleveref} 
\let\labelindent\relax
\usepackage{enumitem}
\usepackage{xspace}

\usepackage{makecell}
\usepackage{multicol}
\usepackage{rotating}
\usepackage{courier}

\usepackage{subcaption} 
\usepackage[font=small]{caption}

\usepackage{pifont}
\definecolor{MyDarkBlue}{rgb}{0,0.08,1}
\definecolor{airforceblue}{rgb}{0.36, 0.54, 0.66}
\definecolor{MyDarkGreen}{rgb}{0.02,0.6,0.02}
\definecolor{MyDarkRed}{rgb}{0.8,0.02,0.02}
\definecolor{MyDarkOrange}{rgb}{0.40,0.2,0.02}
\definecolor{MyPurple}{RGB}{111,0,255}
\definecolor{MyRed}{rgb}{1.0,0.0,0.0}
\definecolor{MyGold}{rgb}{0.75,0.6,0.12}
\definecolor{MyDarkgray}{rgb}{0.66, 0.66, 0.66}
\definecolor{MyPink}{rgb}{0.9, 0.33, 0.5}
\definecolor{MyCyan}{rgb}{0., 0.4, 0.4}
\definecolor{MyBlue}{rgb}{0.5, 0.8, 0.9}
\definecolor{AbsoluteColor}{rgb}{0.76, 0.2, 0.2}
\definecolor{DeltaColor}{rgb}{0.87, 0.72, 0.3}
\definecolor{RelativeColor}{rgb}{0.04, 0.33, 0.58}
\definecolor{StanfordRed}{rgb}{0.549, 0.082, 0.082}

\newcommand{\ours}{UMI}

\newcommand{\mypara}[1]{\par\vspace*{0mm} \textbf{\underline{{#1}}}}

\newcommand\blfootnote[1]{%
  \begingroup
  \renewcommand\thefootnote{}\footnote{#1}%
  \addtocounter{footnote}{-1}%
  \endgroup
}

\begin{document}


\title{\huge{Universal Manipulation Interface:}\\\huge{In-The-Wild Robot Teaching Without In-The-Wild Robots}}


\author{\authorblockN{Cheng Chi$^{* 1, 2}$, Zhenjia Xu$^{* 1,2}$, Chuer Pan$^{1}$, Eric Cousineau$^{3}$, Benjamin Burchfiel$^{3}$, Siyuan Feng$^{3}$, \\
Russ Tedrake$^{3}$, Shuran Song$^{1,2}$} 
${}^{1}$Stanford University, ${}^{2}$ Columbia University, ${}^{3}$Toyota Research Insititute\\
\url{https://umi-gripper.github.io}
}



%


\IEEEpeerreviewmaketitle
\twocolumn[{%
	\renewcommand\twocolumn[1][]{#1}%
	\maketitle
        \vspace{-4mm}
	\begin{center}
		\includegraphics[width=\textwidth]{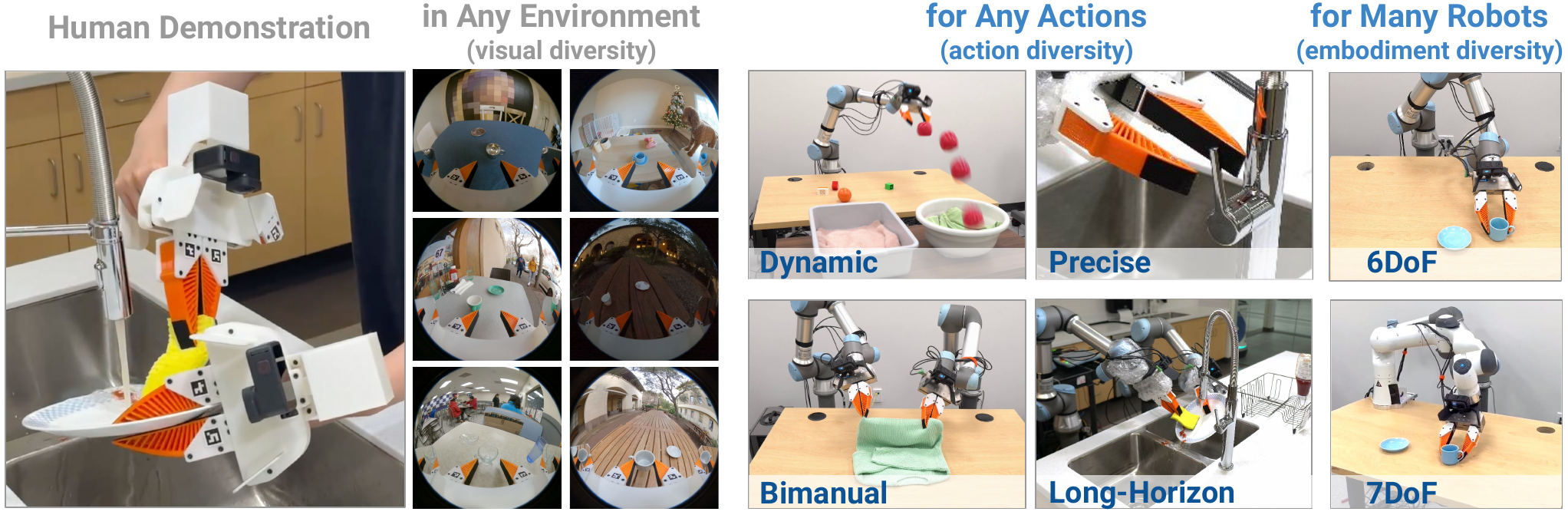}
		\captionof{figure}{\textbf{Universal Manipulation Interface (UMI)} is a portable, intuitive, low-cost data collection and policy learning framework.  This framework allows us to transfer diverse human demonstrations to effective visuomotor policies.  We showcase the framework for tasks that would be difficult with traditional teleoperation, such as dynamic, precise, bimanual and long-horizon tasks. }
  \label{fig:teaser}
	\end{center}
}]

\blfootnote{$*$ Indicates equal contribution}


\begin{abstract}
We present Universal Manipulation Interface (UMI) -- a data collection and policy learning framework that allows direct skill transfer from in-the-wild human demonstrations to deployable robot policies. UMI employs hand-held grippers coupled with careful interface design to enable portable, low-cost, and information-rich data collection for challenging bimanual and dynamic manipulation demonstrations. To facilitate deployable policy learning, UMI incorporates a carefully designed policy interface with inference-time latency matching and a relative-trajectory action representation. The resulting learned policies are hardware-agnostic and deployable across multiple robot platforms. Equipped with these features, UMI framework unlocks new robot manipulation capabilities, allowing zero-shot generalizable dynamic, bimanual, precise, and long-horizon behaviors, by only changing the training data for each task. We demonstrate UMI's versatility and efficacy with comprehensive real-world experiments, where policies learned via UMI zero-shot generalize to novel environments and objects when trained on diverse human demonstrations. UMI's hardware and software system is open-sourced at \href{https://umi-gripper.github.io}{https://umi-gripper.github.io}.
\end{abstract}

\section{Introduction}

How should we demonstrate complex manipulation skills for robots to learn from? 
Attempts in the field have approached this question primarily from two directions: collecting targeted in-the-lab robot datasets via teleoperation or leveraging unstructured in-the-wild human videos. Unfortunately, neither is sufficient, as teleoperation requires high setup costs for hardware and expert operators, while human videos exhibit a large embodiment gap to robots. 

Recently, using sensorized hand-held grippers as a data collection interface \cite{song2020grasping,young2021visual,shafiullah2023dobbe} has emerged as a promising middle-ground alternative -- simultaneously minimizing the embodiment gap while remaining intuitive and flexible. 
Despite their potential, these approaches still struggle to balance action diversity with transferability. While users can theoretically collect any actions with these hand-held devices, much of that data can not be transferred to an effective robot policy.  As a result, despite achieving impressive \textbf{visual diversity} across hundreds of environments, the collected actions are constrained to simple grasping \cite{song2020grasping} or quasi-static pick-and-place \cite{young2021visual,shafiullah2023dobbe}, lacking \textbf{action diversity}.

\begin{figure*}[t] 
    \centering
    \includegraphics[width=\linewidth]{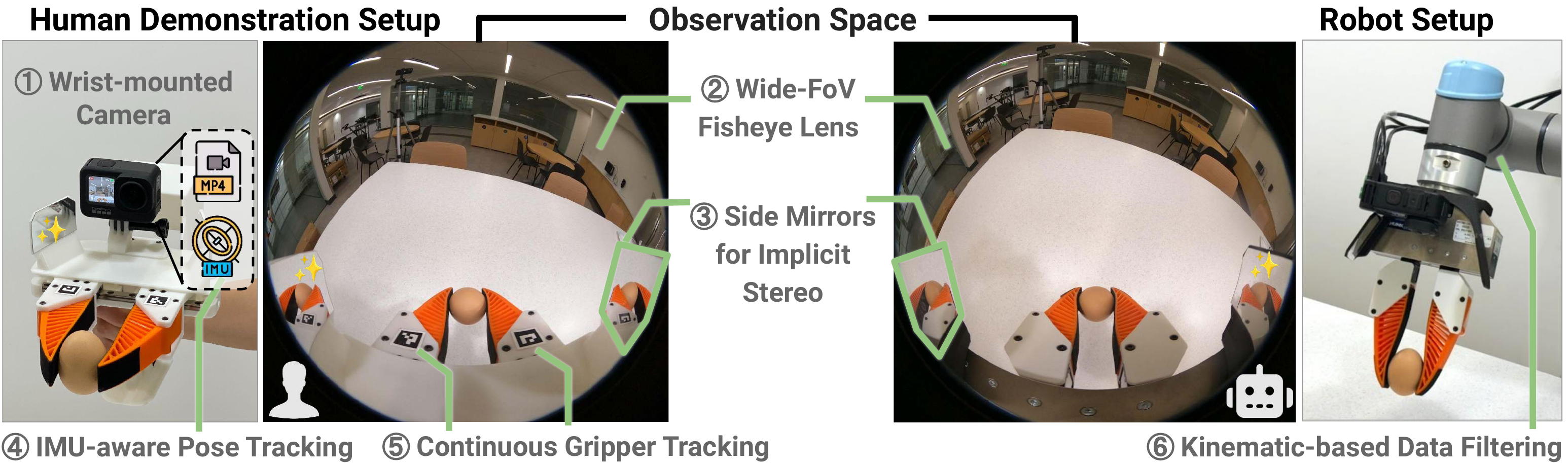}
    \caption{\textbf{UMI Demonstration Interface Design.} 
    Left: Hand-held grippers for data collection, with a GoPro as the only sensor and recording device.
    Middle: Image from the GoPro's 155\degree~Fisheye view. Note the physical side mirrors highlighted in \textcolor{MyDarkGreen}{green} which provide implicit stereo information.
    Right: UMI-compatible robot gripper and camera setup make observation similar to hand-held gripper view.}
    \label{fig:hardware}
    \vspace{-4mm}
\end{figure*}


What prevents action transfer in previous work? 
We identified a few subtle yet critical issues:
\begin{itemize}
    \item \textbf{Insufficient visual context:}  
    While using a wrist-mounted camera is key for aligning the observation space and enhancing device portability, it restricts the scene’s visual coverage. The camera’s proximity to the manipulated object often results in heavy occlusions, providing insufficient visual context for action planning.
    
    

    \item  \textbf{Action imprecision:}
    Most hand-held devices rely on monocular structure-from-motion (SfM) to recover robot actions. However, such methods often struggle to recover precise global action due to scale ambiguity, motion blur, or insufficient texture, which significantly restrict the precision of tasks for which the system can be employed.
    
    \item \textbf{Latency discrepancies:} 
    During hand-held data collection, observation and action recording occur without latency. However, during inference, various latency sources, including sensor, inference, and execution latencies, arise within the system.
    Policies unaware of these latency discrepancies will encounter out-of-distribution input and in turn, generate out-of-sync actions. This issue is especially salient for fast and dynamic actions.


    \item \textbf{Insufficient policy representation:}
    Prior works often use simple policy representations (e.g., MLPs) with action regression loss, limiting their capacity to capture complex multimodal action distributions inherent in human data. 
    Consequently, even with precisely recovered demonstrated actions and all discrepancies removed, the resulting policy could still struggle to fit the data accurately. 
    This further hampers large-scale, distributed human data collection, as more demonstrators increase action multimodality.


\end{itemize}

In this paper, we address these issues with careful design of the demonstration and policy interface: 
\begin{itemize}
\item First, we aim to identify the right \hyperref[sec:hardware_interface]{\textbf{physical interface}} for human demonstration that is intuitive and meanwhile able to capture all the information necessary for policy learning. 
Specifically, we use a {Fisheye lens} to increase the field of view and visual context, and add {side mirrors} on the gripper to provide implicit stereo observation. When combined with the GoPro's built-in {IMU sensor}, we can enable robust tracking under fast motion. 

\item Second, we explore the right \hyperref[sec:policy_interface]{\textbf{policy interface}} (i.e., observation and action representations) that could make the policy hardware-agnostic and thereby enable effective skill transfer.
Concretely, we employ {inference-time latency matching} to handle different sensor observation and execution latency, use relative trajectory as action representation to remove the need for precise global action, and finally, apply Diffusion Policy~\cite{chi2023diffusionpolicy} to model multimodal action distributions. 
\end{itemize}

The final system, \textbf{Universal Manipulation Interface (UMI)}, provides a practical and accessible framework to unlock new robot manipulation skills, allowing us to demonstrate any actions in any environment while maintaining high transferability from human demonstration to robot policy.

With just a wrist-mounted camera on the hand-held gripper (Fig. \ref{fig:hardware}), we show that UMI is capable of achieving a wide range of manipulation tasks that involve dynamic, bimanual, precise and long-horizon actions by only changing the training data for each task~(Fig. \ref{fig:teaser}). 
Furthermore, when trained with diverse human demonstrations, the final policy exhibits zero-shot generalization to novel environments and objects, achieving a remarkable $70\%$ success rate in out-of-distribution tests, a level of generalizabilty seldomly observed in other behavior cloning frameworks. We open-source the hardware and software system at \href{https://umi-gripper.github.io}{https://umi-gripper.github.io}.



\vspace{-2mm}
\section{Related Works}
A key enabler for any data-driven robotics system is the data itself. Here, we review a few typical data collection workflows in the context of robotic manipulation. 

\subsection{Teleoperated Robot Data}
Imitation learning learns policies from expert demonstrations. Behavior cloning (BC), utilizing teleoperated robot demonstrations, stands out for its direct transferability. However, teleoperating real robots for data collection poses significant challenges. Previous approaches utilized interfaces such as 3D spacemouse~\cite{chi2023diffusionpolicy, zhu2022viola}, VR or AR controllers~\cite{seo2023deep, brohan2022rt, ebert2021bridge, jang2022bc, rosete2023latent, zhang2018deep, duan2023ar2}, smartphones~\cite{wang2023mimicplay, wong2022error, mandlekar2018roboturk}, and haptic devices~\cite{shaw2023videodex, wyrobek2008towards, toedtheide2023force, peternel2013learning, brygo2014humanoid} for teleoperation. These methods are either very expensive or hard to use due to high latency and lack of user intuitiveness. While recent advancements in leader-follower (i.e. puppetting) devices such as ALOHA~\cite{zhao2023learning, fu2024mobile} and GELLO~\cite{wu2023gello} offer promise with intuitive and low-cost interfaces, their reliance on real robots during data collection limits the type and number of environments the system can gain access to for ``in-the-wild'' data acquisition. Exoskeletons~\cite{fang2023low, kim2023giving} remove the dependence on real robots during data collection, however, require fine-tuning using teleoperated real robot data for deployment. Moreover, the resulting data and policy from aforementioned devices are embodiment-specific, preventing reusage for different robots. 

In contrast, \ours~eliminates the need for physical robots during data collection and offers a more portable interface for in-the-wild robot teaching, providing data and policies that are transferable to different robot embodiments (e.g., 6DoF or 7DoF robot arms). 



\subsection{Visual Demonstrations from Human Video}
There's a distinct line of work dedicated to policy learning from in-the-wild video data (e.g. YouTube videos). The most common way is to learn from diverse passive human demonstration videos. Utilizing passive human demonstrations, previous works learn task cost functions~\cite{shao2021concept2robot, chen2021learning, bahl2022human, ma2022vip}, affordance functions~\cite{bahl2023affordances}, dense object descriptors~\cite{simeonov2022neural, pmlr-v205-pan23a, shen2023distilled}, action correspondences~\cite{schmeckpeper2020learning, qin2022dexmv}, and pre-trained visual representations~\cite{nair2022r3m, xiao2022masked}.

However, this approach encounters three major challenges. Firstly, most video demonstrations lack explicit action information, crucial for learning generalizable policies. To infer action data from passive human video, previous works resort to hand pose detectors~\cite{wang2023mimicplay, bahl2022human, shaw2023videodex, qin2022dexmv}, or combining human videos with in-domain teleoperated robot data to predict actions~\cite{schmeckpeper2020learning, kim2023giving, schmeckpeper2020reinforcement, qin2022dexmv}. Second, the evident embodiment gap between humans and robots hinders action transfer. Efforts to bridge the gap include learning human-to-robot action mapping with hand pose retargetting~\cite{shaw2023videodex, qin2022dexmv} or extracting embodiment-agnostic keypoints~\cite{xiong2021learning}. Despite these attempts, the inherent embodiment differences still complicate policy transfer from human video to physical robots. Thirdly, the inherent observation gap induced by the embodiment gap in this line of work introduces inevitable mismatch between train/inference time observation data, exacerbating the transferability of the resulting policies, despite efforts in aligning demonstration observation with robot observation ~\cite{kim2023giving, qin2022dexmv}.

In contrast, data collected with \ours~exhibit minimal embodiment gap both in action and observation spaces, enabled by precise manipulation action extraction via robust visual-inertial camera tracking and the shared Fisheye wrist-mounted cameras during teaching and testing. 
Consequently, this enables in-the-wild zero-shot policy transfer for dynamic, bimanual, precise, and long-horizon manipulation tasks.

\vspace{-1mm}
\subsection{Hand-Held Grippers for Quasi-static Actions}

Hand-held grippers~\cite{song2020grasping, young2021visual, doshi2023hand, sanches2023scalable, praveena2019characterizing, pari2021surprising} minimize observation embodiment gaps in manipulation data collection, offering portability and intuitive interfaces for efficient data collection in the wild. However, accurately and robustly extracting 6DoF end-effector (EE) pose from these devices remains challenging, hindering the deployment of robot policies learned from these data on fine-grained manipulation tasks.

Prior works attempted to address this issue through various approaches, such as 
SfM~\cite{young2021visual, pari2021surprising} which suffers from scale ambiguity; 
RGB-D fusion~\cite{song2020grasping} which requires expensive sensors and onboard compute;
external motion tracking~\cite{sanches2023scalable, praveena2019characterizing} which is limited to lab settings.
These devices, constrained to quasi-static actions due to low EE tracking accuracy and robustness, often necessitate cumbersome onboard computer or external motion capture (MoCap) systems, diminishing their feasibility for in-the-wild data collection. In contrast, \ours~integrates state-of-the-art SLAM~\cite{campos2021orb} with built-in IMU data from GoPro, to accurately capture 6DoF actions at the global scale. The high-accuracy data enables trained BC policy to learn bimanual tasks. With thorough latency matching, \ours~further enables real-world deployable policy for dynamic actions such as tossing.

Recently, Dobb-E ~\cite{shafiullah2023dobbe} proposed a ``reacher-grabber'' tool mounted with an iPhone 
to collect single-arm demonstrations for the Stretch robot.
Yet, Dobb-E only demonstrates policy deployment for quasi-static tasks and requires environment-specific policy fine-tuning. Conversely, using only data collected with \ours~enables trained policy to zero-shot generalize to novel in-the-wild environments, unseen objects, multiple robot embodiments, for dynamic, bimanual, precise and long-horizon tasks.

\begin{figure}[t]
    \centering
    \includegraphics[width=0.98\linewidth]{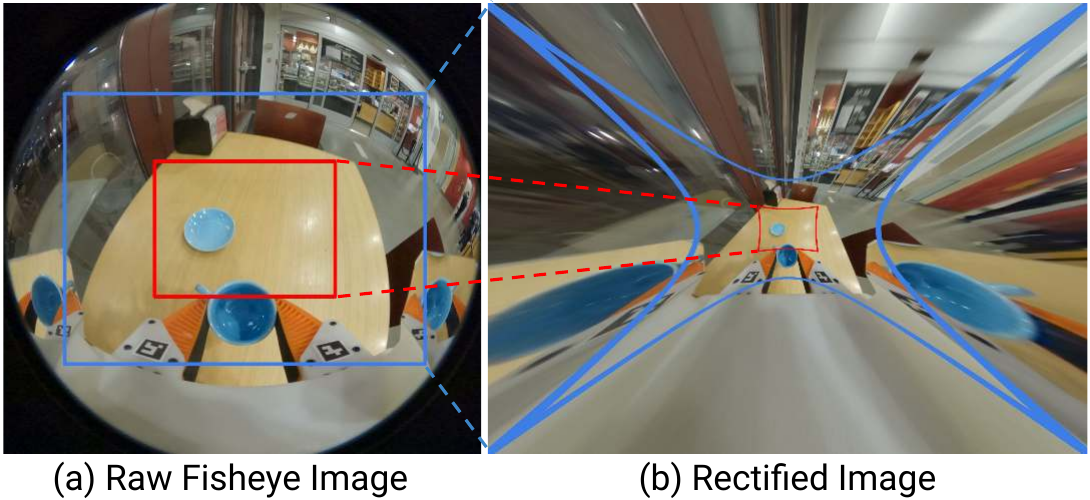}
    \caption{\textbf{Fisheye vs Rectilinear}
    (a) UMI policies use raw Fisheye image as observation. 
    (b) Rectifying a large 155\degree~FoV image to the pin-hole model severely stretches the peripheral view (outside of \textcolor{blue}{blue line}), while compresses the most important information at the center to a small area (inside of \textcolor{red}{red line}).} 
    \vspace{-7mm}
    \label{fig:fisheye}
\end{figure}

\vspace{-1mm}
\section{Method}
Universal Manipulation Interface (UMI) is hand-held data collection and policy learning framework 
that allows direct transfer from in-the-wild human demonstrations to deployable robot policies. It is designed with the following goals in mind: 
\begin{itemize}[leftmargin=4mm] 
    \item \textbf{Portable.} The hand-held UMI grippers can be taken to any environment and start data collection with close-to-zero setup time.
    \item \textbf{Capable.} The ability to capture and transfer natural and complex human manipulation skills beyond pick-and-place. 
    \item \textbf{Sufficient.} The collected data should contain sufficient information for learning effective robot policies and contain minimal embodiment-specific information that would prevent transfer.
    \item \textbf{Reproducible}: Researchers and enthusiasts should be able to consistently build UMI grippers and use data to train their own robots, even with different robot arms.
\end{itemize}

The following sections describe how we enable the above goals through our hardware and policy interface design. 

\subsection{Demonstration Interface Design}
\label{sec:hardware_interface}
UMI's data collection hardware takes the form of a trigger-activated, handheld 3D printed parallel jaw gripper with soft fingers, mounted with a GoPro camera as the \textbf{only} sensor and recording device (see \hyperref[HD1]{HD1}). For bimanual manipulation, UMI can be trivially extended with another gripper. The key research question we need to address here is: 
\begin{center}
\emph{How can we capture sufficient information for a wide \\variety of tasks with just a wrist-mounted camera?}
\end{center}

Specifically, on the \textbf{observation} side, the device needs to capture sufficient visual context to infer action \hyperref[HD2]{HD2} and critical depth information  \hyperref[HD3]{HD3}. 
On the \textbf{action} side, it needs to capture precise robot action under fast human motion \hyperref[HD4]{HD4}, detailed subtle adjustments on griping width \hyperref[HD5]{HD5}, and automatically check whether each demonstration is valid given the robot hardware kinematics \hyperref[HD6]{HD6}. 
The following sections describe details on how we achieve these goals. 



\textbf{HD1. Wrist-mounted cameras as input observation.} 
\label{HD1}
We rely solely on wrist-mounted cameras, without the need for any external camera setups. When deploying UMI on a robot, we place GoPro cameras with the same location with respect to the same 3D-printed fingers as on the hand-held gripper. This design provides the following benefits: 

\begin{enumerate}[leftmargin=4mm]
    \item \textbf{Minimizing the observation embodiment gaps.} Thanks to our hardware design, the videos observed in wrist-mount cameras are almost indistinguishable between human demonstrations and robot deployment, making the policy input less sensitive to embodiment.

    \item  \textbf{Mechanical robustness.} 
    Because the camera is mechanically fixed relative to the fingers, mounting UMI on robots does not require camera-robot-world calibration. Hence, the system is much more robust to mechanical shocks, making it easy to deploy.

    \item \textbf{Portable hardware setup.} Without the need for an external static camera or additional onboard compute, we largely simplify the data collection setup and make the whole system highly portable.

    \item \textbf{Camera motion for natural data diversification.} A side benefit we observed from experiments is that when training with a moving camera, the policy learns to focus on task-relevant objects or regions instead of background structures (similar in effect to random cropping). As a result, the final policy naturally becomes more robust against distractors at inference time. 
    
\end{enumerate}

 Avoiding use of external static cameras also introduce additional challenges for downstream policy learning. For example, the policy now needs to handle non-stationary and partial observations. We mitigated these issues by leveraging wide-FoV Fisheye Lens \hyperref[HD2]{HD2}, and robust visual tracking \hyperref[HD4]{HD4}, described in the following sections.

\textbf{HD2. Fisheye Lens for visual context.}
\label{HD2}
We use a 155-degree Fisheye lens attachment on wrist-mounted GoPro camera, which provides sufficient visual context for a wide range of tasks, as shown in Fig. \ref{fig:hardware}. As the policy input, we directly use raw Fisheye images \textit{without undistortion} since Fisheye effects conveniently preserve resolution in the center while compressing information in the peripheral view. In contrast, rectified pinhole image (Fig. \ref{fig:fisheye} right) exhibits extreme distortions, making it unsuitable for learning due to the wide FoV.
Beyond improving SLAM robustness with increased visual features and overlap \cite{zhang2016fov}, our quantitative evaluation (Sec \ref{sec:eval_cup_no_fisheye}) shows that the Fisheye lens improves policy performance by providing the necessary visual context.


\begin{figure}[t]
    \centering
    \includegraphics[width=0.95\linewidth]{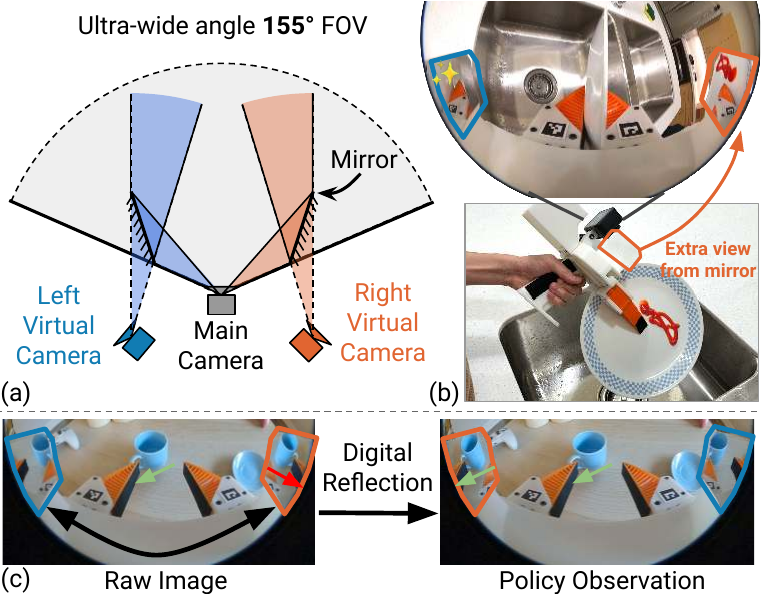}
    \caption{
    \textbf{UMI Side Mirrors.} The ultra-wide-angle camera coupled with strategically positioned mirrors, facilitates implicit stereo depth estimation. 
    \textbf{(a)}: The view through each mirror effectively creates two virtual cameras, whose poses are reflected along the mirror planes with respect to the main camera.
    \textbf{(b)}: Ketchup on the plate, occluded from the main camera view, is visible inside the right mirror, proving that mirrors simulate cameras with different optical centers.
    \textbf{(c)}: We digitally reflect the content inside mirrors for policy observation. Note the orientation of the cup handle becomes consistent across all 3 views after reflection.
    }
    \vspace{-7mm}
    \label{fig:mirror}
\end{figure}

\begin{figure*}[t]
    \centering
    \includegraphics[width=\linewidth]{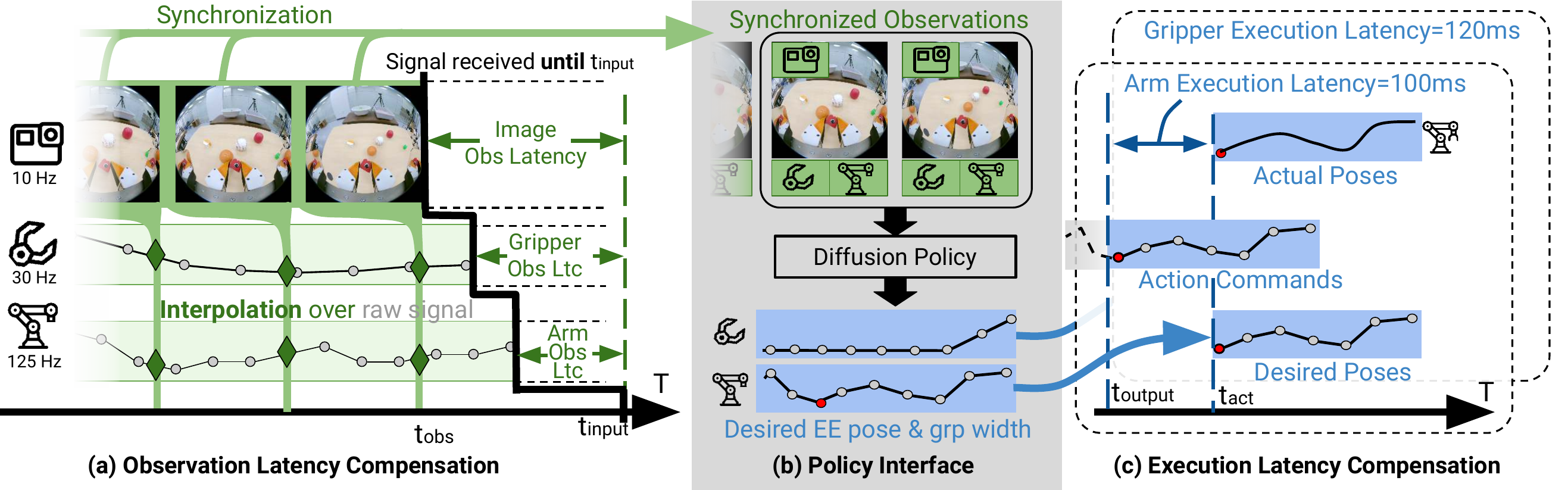}
    \caption{
    \textbf{\textbf{UMI Policy Interface Design.}}
    (b) UMI policy takes in a sequence of \textbf{synchronized} observations (RGB image, relative EE pose, and gripper width) and outputs a sequence of \textbf{desired} relative EE pose and gripper width as action.
    (a) We synchronize different observation streams with physically measured latencies.
    (c) We send action commands ahead of time to compensate for robots' execution latency.
    }
    \label{fig:method}
    \vspace{-4mm}
\end{figure*}

\textbf{HD3. Side mirrors for implicit stereo.} 
\label{HD3}
To mitigate the lack of direct depth perception from the monocular camera view, we placed a pair of physical mirrors in the cameras' peripheral view which creates implicit stereo views all in the same image. As illustrated in Fig \ref{fig:mirror} (a), the images inside the mirrors are equivalent to what can be seen from additional cameras reflected along the mirror plane, without the additional cost and weight. To make use of these mirror views, we found that digitally reflecting the crop of the images in the mirrors, shown in Fig \ref{fig:mirror} (c), yields the best result for policy learning (Sec. \ref{sec:eval_cup_mirror}). Note that without digital reflection, the orientation of objects seen through side mirrors is the opposite of that in the main camera view.

\textbf{HD4. IMU-aware tracking.} 
\label{HD4}
UMI captures rapid movements with absolute scale by leveraging GoPro's built-in capability to record IMU data (accelerometer and gyroscope) into standard mp4 video files \cite{gpmf}.
By jointly optimizing visual tracking and inertial pose constraints, our Inertial-monocular SLAM system based on ORB-SLAM3 \cite{orbslam3} maintains tracking for a short period of time even if visual tracking fails due to motion blur or a lack of visual features (e.g. looking down at a table). This allows UMI to capture and deploy highly dynamic actions such as tossing (shown in Fig \ref{fig:task}). 
In addition, the joint visual-inertial optimization allows direct recovery of real metric scale, important for action precision and inter-gripper pose proprioception \hyperref[PD2.3]{PD2.3}: a critical ingredient to enable bimanual policy.


\textbf{HD5. Continuous gripper control.} 
\label{HD5}
In contrast to the binary open-close action used in prior works \cite{song2020grasping, wang2023mimicplay, zhu2022viola}, we found commanding gripper width continuously significantly expands the range of tasks doable by parallel-jaw grippers. For example, the tossing task (Fig. \ref{fig:task}) requires precise timing for releasing objects. Since objects have different widths, binary gripper actions will be unlikely to meet the precision requirement. 
On UMI gripper, finger width is continuously tracked via fiducial markers \cite{aruco} (Fig. \ref{fig:hardware} left). 
Using series-elastic end effectors principle \cite{suh2022seed}, UMI can implicitly record and control grasp forces by regulating the deformation of soft fingers through continuous gripper width control.

\textbf{HD6. Kinematic-based data filtering.}
\label{HD6}
While the data collection process is robot-agnostic, we apply simple kinematic-based data filtering to select valid trajectories for different robot embodiments.   
Concretely, when the robot's base location and kinematics are known, the absolute end-effector pose recovered by SLAM allows kinematics and dynamics feasibility filtering on the demonstration data. Training on the filtered dataset ensures policies comply with embodiment-specific kinematic constraints. 


\textbf{Putting everything together.} The UMI gripper weighs 780g, with an external dimension of $L310mm \times W175mm \times H210mm$ and finger stroke of $80mm$. The 3D printed gripper has a BoM cost of \textbf{\$73}, while the GoPro camera and accessories total \textbf{\$298}. As shown in Fig. \ref{fig:hardware}, we can equip any robot arms with a compatible gripper and camera setup.

\subsection{Policy Interface Design}
\label{sec:policy_interface}

With the collected demonstration data, we can train a visuomotor policy that takes in a sequence of synchronized observations (RGB images, 6 degrees-of-freedom end-effector pose, and gripper width) and produces a sequence of actions (end-effector pose and gripper width) as shown in Fig. \ref{fig:method} (b). 
In this paper, we use Diffusion Policy \cite{chi2023diffusionpolicy} for all of our experiments, while other frameworks such as ACT \cite{zhao2023learning} could potentially serve as a drop-in replacement.

An important goal of UMI's policy interface design is to ensure the interface is \textbf{agnostic to underlying robotic hardware platforms} 
such that the resulting policy, trained on one data source (i.e., hand-held gripper), could be directly deployed to different robot platforms. 
To do so, we aim to address the following two key challenges: 
\begin{itemize}
    \item \textbf{Hardware-specific latency.} The latency of various hardware (streaming camera, robot controller, industrial gripper) is highly variable across system deployments, ranging from single-digit to hundreds of milliseconds. In contrast, all information streams captured by UMI grippers have zero latency with respect to the image observation, thanks to GoPro's synchronized video, IMU measurements and the vision-based gripper width estimation.
    \item \textbf{Embodiment-specific proprioception.} Commonly used proprioception observations such as joint angles and EE pose are only well-defined with respect to a specific robot arm and robot base placement. In contrast, UMI needs to collect data across diverse environments and be generalizable to multiple robot embodiments.
\end{itemize}
In the following sections, we will describe three policy interface designs that address these challenges.

\textbf{PD1. Inference-time latency matching.} 
\label{PD1}
While UMI's policy interface assumes synchronized observation streams and immediate action execution, physical robot systems do not conform to this assumption. If not carefully handled, the timing mismatch between training and testing can cause large performance drops on dynamic manipulation tasks that require rapid movement and precise hand-eye coordination, demonstrated in Sec \ref{sec:eval_toss_latency}. 
In this paper, we separately handle timing discrepancies on the observation and action sides: 

\textbf{PD1.1) Observation latency matching.}
On real robotic systems, different observation streams (RGB image, EE pose, gripper width) are captured by distributed micro-controllers, resulting in different observation latency. 

%

For each observation stream, we individually measure their latency (details see \S \ref{sec:camera_latency}-\ref{sec:gripper_latency}).
At inference time, we align all observations with respect to the stream with the highest latency (usually the camera). Specifically, we first temporally down-sample the RGB camera observations to the desired frequency (often 10-20Hz), and then use the {capture timestamp} of each image $t_{obs}$ to linearly interpolate gripper and robot proprioception streams. 
In bimanual systems, we soft-synchronize two cameras by finding the nearest neighbor frames, which can be off by a maximum of $\frac{1}{60}$ seconds. The result is a sequence of synchronized observations that conform to UMI policy, shown in Fig. \ref{fig:method} (a).

\textbf{PD1.2) Action latency matching.}
UMI policy assumes the output as a sequence of synchronized EE poses and gripper widths. However, in practice, robot arms and grippers can only track the desired pose sequence up to an \textbf{execution latency}, that varies across different robot hardware. To make sure the robots and grippers reach the desired pose at the desired time (given by the policy), we need to send commands \textbf{ahead of time} to compensate for execution latency, as shown in Fig. \ref{fig:method} (c). See \S \ref{sec:execution_latency} 
 for execution latency calibration details.

During execution, the UMI policy predicts the action sequence starting at the last step of observation $t_{obs}$. The first few actions predicted are immediately outdated due to observation latency $t_{input}-t_{obs}$, policy inference latency $t_{output}-t_{input}$ and execution latency $t_{act} - t_{output}$. We simply discard the outdated actions and only execute actions with the desired timestamp after $t_{act}$ for each hardware.

\textbf{PD2. Relative end-effector pose.}
\label{PD2}
End-effector (EE) pose is central to both UMI's observation and action space. To avoid dependence on embodiment/deployment-specific coordinates, we represent all EE poses relative to gripper's current EE pose.

\textbf{PD2.1) Relative EE trajectory as action representation.}
\label{PD2.1} Prior works have shown the significant impact of action space selection on task performance \cite{chi2023diffusionpolicy}, with experimental evidence favoring absolute positional actions over delta actions. However, we found that a relative trajectory representation, defined for an action sequence starting at $t_0$ as a sequence of $SE(3)$ transforms denoting the desired pose at $t$ relative to the initial EE pose at $t_0$,  allows the system to be more robust against tracking errors during data collection and camera displacements.  


\textbf{PD2.2) Relative EE trajectory as proprioception.}
\label{PD2.2}
Similarly, we represent the proprioception of history EE poses as a relative trajectory. When observation horizon is set to 2, this representation effectively provides velocity information to the policy. Combined with our wrist-mounted camera observation space, relative trajectory allows our system to be \textbf{calibration-free}. Moving the robot base during execution will not affect task performance (Fig. \ref{fig:UMI_robustness} (a)), as long as the objects are still within reach range, making the UMI framework applicable to mobile manipulators as well. 

\begin{figure}[ht]
    \centering
    \includegraphics[width=0.85\linewidth]{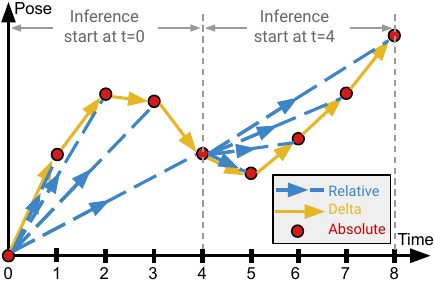}
    \caption{
    \textbf{Relative Trajectory as Action Representation}. 
    \textcolor{RelativeColor}{Relative trajectory}, used by UMI, is a sequence of end-effector (EE) poses relative to \textbf{the same current EE pose} for each inference step. In contrast, \textcolor{DeltaColor}{Delta action} represents each action step relative to its immediate previous action, therefore accumulates error. \textcolor{AbsoluteColor}{Absolute action} requires a global coordinate frame for all actions, which is difficult to define for in-the-wild data collection.
   } \vspace{-2mm}
    \label{fig:action_repr}
\end{figure}

\textbf{PD2.3) Relative inter-gripper proprioception.}
\label{PD2.3}
When using UMI in a bimanual setup, we found that providing the policy with the relative pose between the two grippers to be critical for bimanual coordination and task success, as shown in Sec. \ref{sec:eval_cloth}. The effect of inter-gripper proprioception is particularly large when the visual overlap between two cameras is small. 
The inter-gripper proprioception is enabled by our map-then-localize data collection scheme that constructs a scene-level global coordinate system \hyperref[HD4]{HD4}.
For each new scene, we first collect a video that builds a map for the scene.
Then, all demonstrations collected in this scene are relocalized to the same map, therefore sharing the same coordinate system. Despite the videos from each gripper being relocalized separately, the relative pose between two grippers at each time step can be calculated using their shared coordinates.





\section{Evaluations}

\begin{figure*}[t] 
    \centering
    \includegraphics[width=\linewidth]{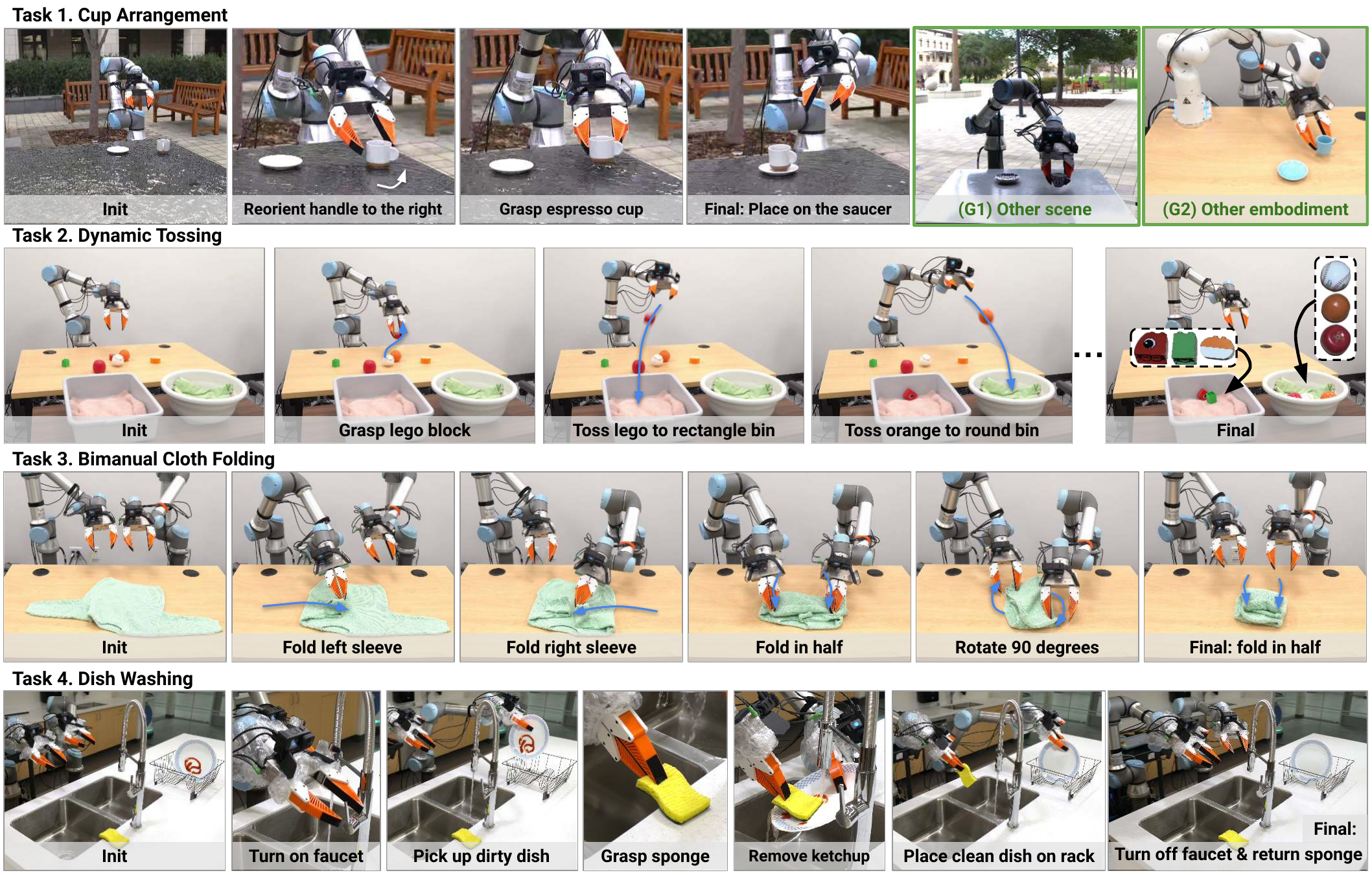}
    \caption{
    \textbf{Policy Rollouts.} We test UMI on a variety of challenging real-world tasks. 
     \textbf{Cup arrangement} tests UMI's ability to learn both prehensile and non-prehensile actions, and to capture multi-modal action distribution (clockwise and counter-clockwise rotation).  This task is evaluated in both narrow-domain and unseen environments as well as two robot embodiments.
    \textbf{Dynamic tossing} tests UMI’s ability to capture and transfer rapid human motions as well as precise hand-eye coordination.
    \textbf{Bimanual cloth folding} tests UMI’s ability to synchronize two-arm coordination. 
    \textbf{Dish washing} tests UMI’s ability to handle long-horizon tasks that involve multiple rigid, deformable, and articulated objects. Please check videos on our \href{https://umi-gripper.github.io}{website} for more details. }
    \label{fig:task}
    \vspace{-4mm}
\end{figure*}
In our experiment, we aim to evaluate the UMI framework's effectiveness for deployable policy learning in three aspects:

\begin{itemize}
    \item \textbf{Capability:} How well can we transfer UMI demonstrations to effective robot policy? Especially for complex, dynamic, bimanual, and long-horizon manipulation skills.
    
    \item \textbf{Generalization:} Will data collected in the wild within diverse environments help the policy to generalize to unseen environments and objects? 

    \item \textbf{Data collection efficiency:} How fast can we collect manipulation data with UMI? What's the accuracy of the SLAM system? 
\end{itemize}

To access capability and generalization, we evaluate UMI on 4 real-world robotic tasks across both narrow domain and in-the-wild environments, shown in Fig. \ref{fig:task}. To measure data collection efficiency, we compare the UMI gripper with human hand demonstration and a typical teleop interface. See \S \ref{sec:supp_data_collection} for detailed data collection protocol.

\section{Capability Experiments}

We study UMI's ability to capture and transfer single-hand, bimanual, dynamic, and long-horizon manipulation skills with four tasks. For capability experiments, all tasks are evaluated in the same environment as data collection but with randomized robot and object initial states. To ensure a fair comparison, we use exactly \textit{the same initial state across all methods} for both the robot and objects, by manually aligning the scene against pre-recorded images. 
See \S \ref{sec:supp_evaluation} for detailed evaluation protocol and videos for all experiments.

\begin{figure*}
\centering
\includegraphics[width=\linewidth]{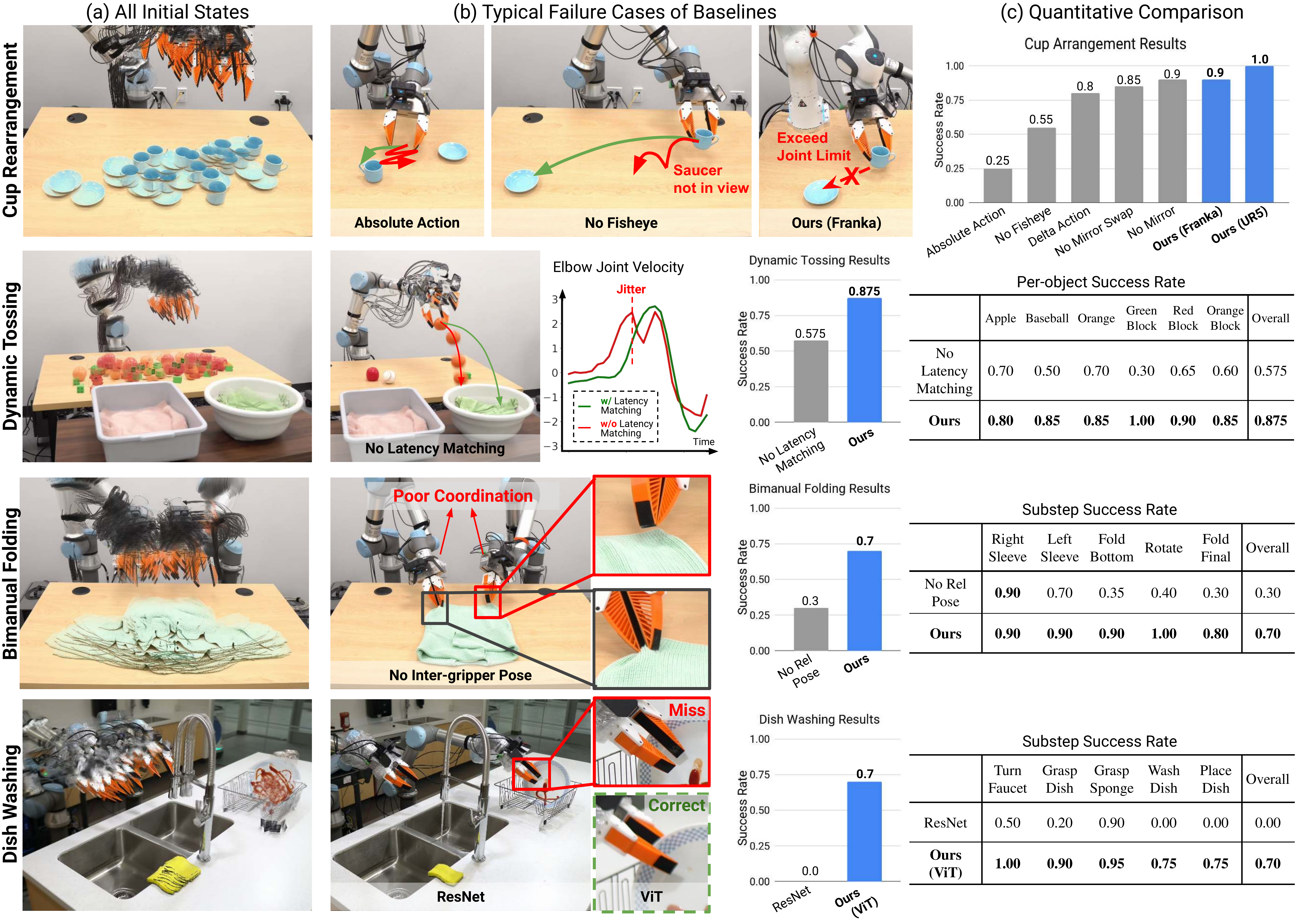}
\caption{
\textbf{Narrow-domain Evaluation Results.}
(a) Initial states for all evaluation episodes overlayed together. For each task, all methods start with the same set of initial states, matched manually with reference images.
(b) Typical failure mode of the baseline/ablation policy. The red arrow indicates failure behavior, green arrow indicates desired behavior. 
(c) Success rate over 20 evaluation episodes, best performance for each column are bolded. Please check our \href{https://umi-gripper.github.io}{website} for more comparison videos. 
\vspace{-4mm}
}
\label{fig:policy_learning}
\end{figure*}

\vspace{-2mm}
\subsection{\textbf{Cup Arrangement}} 
\label{sec:eval_cup}
\mypara{Task} Place an espresso cup on the saucer with its handle facing to the left of the robot, Fig. \ref{fig:task}. 
We defined task success as when the cup is placed upright on the saucer with its handle within $\pm15\degree$ to the left.

\mypara{Capability (what makes the task difficult?)} This task tests the system’s ability to learn both {prehensile} (pick and place) and {non-prehensile} actions (i.e., pushing to reorientate the cup). When the handle faces straight away from the robot, the two equally valid solutions: rotation clockwise and counter-clockwise form a {multi-modal action distribution}.
This task also tests UMI's ability to sense {relative depth} through monocular camera observation and side mirrors.

\mypara{Performance} The training dataset contains 305 episodes collected by 2 demonstrators, evaluation includes 20 test cases, with the testing initial state distribution shown in Fig. \ref{fig:policy_learning} (a). UMI can complete the task 20/20. The next paragraphs will discuss our ablation studies around our key design decisions. 

\textbf{Cross-robot generalization:}
To demonstrate UMI's cross-embodiment generality, we also deployed the same policy checkpoint on a Franka Emika FR2 robot, shown in Fig. \ref{fig:teaser} and Fig. \ref{fig:policy_learning}. This experiment achieves $18/20=90\%$ success rate, with the 2 failure cases being joint limit violations, which could have been avoided if we had mounted the FR2 robot at a different location.

\textbf{No Fisheye lens \hyperref[HD2]{[HD2]}:}
\label{sec:eval_cup_no_fisheye}
To ablate the importance of having a wide field-of-view (FoV) Fisheye lens, we post-processed the dataset by rectifying and cropping each image to a square with $69\degree$ horizontal and vertical FoV. This is a generous analogy of RealSense D415 ($69\degree$ HFoV, $42\degree$ VFoV) and iPhone wide camera ($69\degree$ HFoV, $51\degree$ VFoV).
This baseline only achieves $11/20=55\%$ success rate. Beyond the expected failure mode where the cup is outside of camera view, we found this baseline policy to perform surprisingly poor even if the object is visible, with often jittery motions. We suspect that during training, the poor object visibility forced the policy to be unnecessarily multimodal. 

\textbf{Alternative action spaces \hyperref[PD2]{[PD2]}:}
As alternatives to our relative trajectory as action representation, we also consider absolute and delta action spaces as illustrated in Fig \ref{fig:action_repr}. Since the SLAM system outputs pose relative to the first frame of the mapping video (details in \S \ref{sec:supp_slam}), we can only calculate relative and delta actions directly using SLAM output. To compute absolute actions in the robot base frame, we calibrate both SLAM coordinates and the robot with respect to the same fiducial markers \cite{aruco} placed on the table.

The delta action baseline achieves $16/20=80\%$ success rate. 
The absolute action baseline performs surprisingly poorly with only $5/20=25\%$ success rate, demonstrating a noticeable bias in action selection, likely due to inaccurate calibration between the SLAM and robot base coordinate frames (Fig. \ref{fig:policy_learning} (b)).
While theoretically the performance of this baseline could approach that of relative trajectory with better calibration, this experiment underscores the difficulty of obtaining action data with absolute coordinates, even in controlled lab settings.

\textbf{Effect of side mirrors \hyperref[HD3]{[HD3]}:}
\label{sec:eval_cup_mirror}
To our surprise, directly providing mirror images decreases the performance from $18/20=90\%$ (no mirror) to $17/20=85\%$. 
To fully take advantage of side mirrors, we need to digitally reflect the content inside mirrors and swap left and right mirror images, which achieves a $20/20=100\%$ success rate. We hypothesize that without digital reflection, the opposite motions observed in the main and mirrored images might confuse vision encoders, especially those with translational equivariance.

\begin{figure*}
\centering
\includegraphics[width=\linewidth]{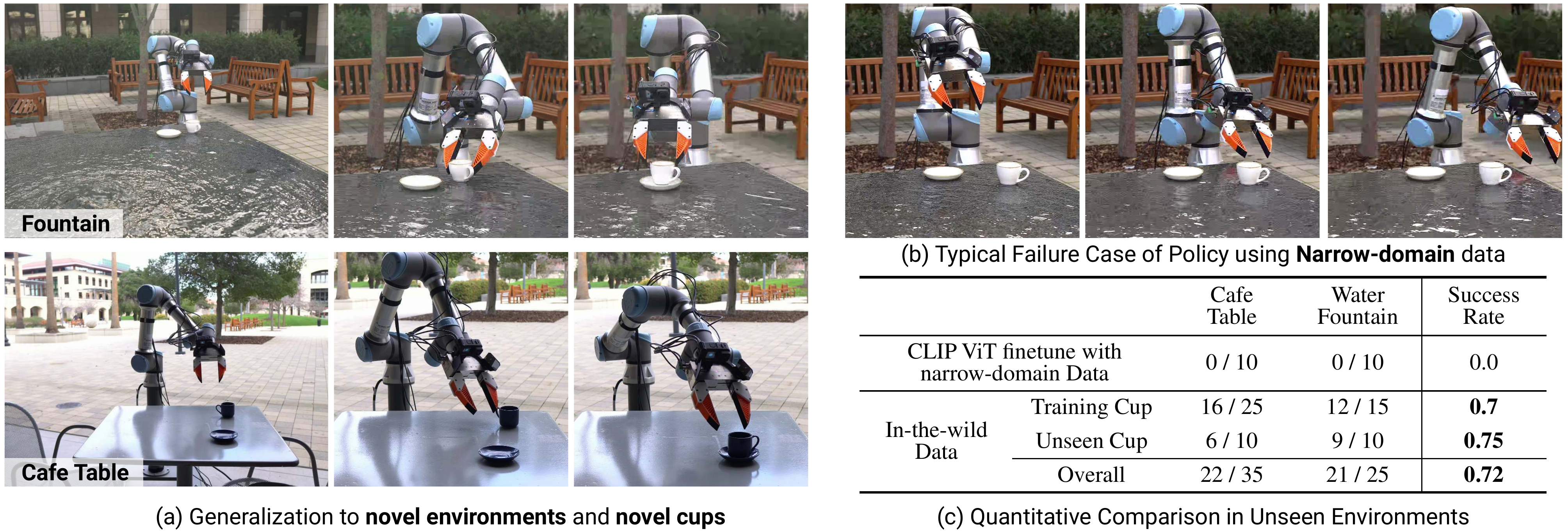}
\caption{
\textbf{In-the-wild Cup Arrangement Evaluation.} (a) The policy, trained with in-the-wild data, demonstrates strong generalization capability to novel environments and novel objects. (b) With only narrow-domain data, the policy struggles to generalize to new environments. (c) Comprehensive qualitative results are provided, and additional comparison videos can be found on our \href{https://umi-gripper.github.io}{website}.
}
\label{fig:in_the_wild_evaluation}
\vspace{-4mm}
\end{figure*}

\subsection{\textbf{Dynamic Tossing}}
\mypara{Task} The robot is tasked to sort 6 objects from the YCB object set \cite{ycb} randomly placed on a table by tossing them to the corresponding bin. The 3 spherical objects (baseball, orange, apple) should be tossed into the round bin, while the 3 Lego Duplo pieces go into the rectangular bin (Fig. \ref{fig:task}). The bins are placed beyond the robot's kinematic reach range to highlight the necessity of dynamic action for this task. 

\mypara{Capability:} The dynamic tossing task demonstrates UMI's ability to capture and transfer fluid and rapid human motions, precise hand-eye coordination (between RGB and proprioception) and timing alignment (between robot and gripper).

\mypara{Performance:} We collected 280 demonstration episodes for this task, with mixed multi and single-object picking and tossing.
Our policy (with inference time latency matching) achieves $105/120=87.5\%$ success rate, counted by the number of objects successfully tossed to their corresponding bin.

\textbf{No Latency Matching \hyperref[PD1]{[PD1]}:}
\label{sec:eval_toss_latency}
With the same trained policy, we disable inference-time latency matching by setting the measured latencies for all observation and action streams to 0.
We visually observe the policy's movement is much more jittery due to the out-of-sync observations and executions. 

While the jitteriness minimally affects grasping, its impact on tossing performance is notable as it disrupts the robot motion to achieve the desired tossing velocity, as illustrated in the elbow joint velocity curve in Fig. \ref{fig:policy_learning}.
In addition, the misalignment between the gripper and robot action (due to their different execution latency) leads to suboptimal object release during tossing. As a result, the final success rate decreased to $69/120=57.5\%$.

%


\subsection{\textbf{Bimanual Cloth Folding}}
\label{sec:eval_cloth}

\mypara{Task} Two robot arms need to coordinate and fold the sweater's sleeves inward, fold up the bottom hem, rotate 90 degrees, and finally fold the sweater in half again (Fig. \ref{fig:task}). See \S \ref{sec:supp_evaluation} for detailed evaluation protocol.

\mypara{Capability} Manipulating high degrees of freedom deformable objects is challenging due to their complex dynamics and underactuation. 
In addition, this task requires tight coordination between arms.  For example, lifting the bottom hem requires two arms to pick it up at the same time, and if one of the arm is just slightly too fast or slow this step will fail.

\mypara{Performance} We collected 250 demonstrations from two demonstrators, with randomized initial states by translating/rotating the shirt and folding the sleeves.
We use a single centralized policy to generate actions for both robot arms and grippers.
Our policy achieves $14/20=70\%$ success rate. 

\textbf{No relative inter-gripper proprioception \hyperref[PD2.3]{[PD2.3]}: }
Without inter-gripper proprioception information (during both training and eval), the coordination between the two arms becomes significantly worse. 
The most salient failure case is when the two arms lift the bottom hem of the shirt, where the baseline policy often misses one of the grasps due to asynchronous grasp action (Fig. \ref{fig:policy_learning} (b)).  
As a result, the baseline policy only achieves success rate of $6/20=30\%$.
In contrast, UMI policy synchronizes the grasp by first reaching the pre-grasp pose and waiting until both arms are in position before simultaneously grasping and folding.


\begin{figure*}[t]
\centering
\includegraphics[width=\linewidth]{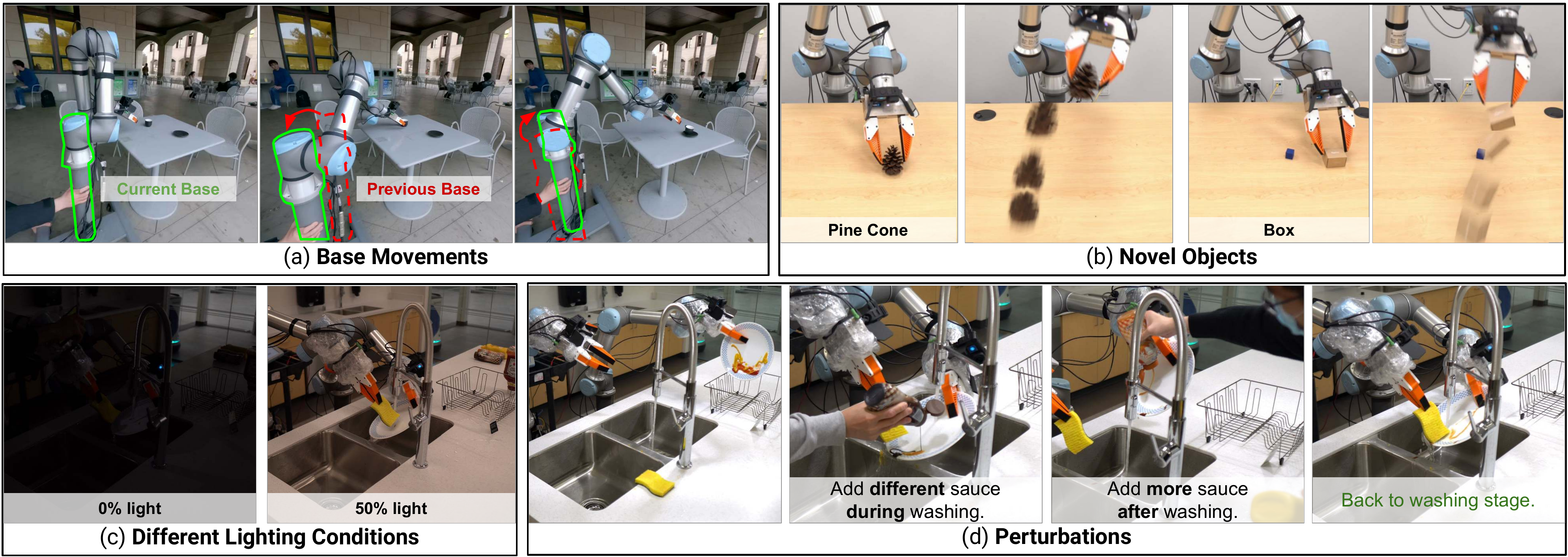}

\caption{\textbf{Robustness Test.} We test the policy robustness with different inference time 
 perturbations such as moving robot base, novel objects, different lighting conditions, and adding different and more condiments for the dish-washing task. The learned policy is robust against these perturbations and completes the task. Please check the video on our \href{https://umi-gripper.github.io}{website} for more details. 
}
\label{fig:UMI_robustness}
\end{figure*}

\subsection{\textbf{Dish Washing}}
\label{sec:eval_dish}
\mypara{Task} The robot needs to execute \textbf{7 steps} of sequentially dependent actions (turn on faucet, grasp plate, pick up sponge, wash and wipe plate until ketchups are removed, place plate, place the sponge and turn off faucet), shown in Fig. \ref{fig:task}. See \S \ref{sec:supp_evaluation} for detailed evaluation protocol.

\mypara{Capability} This task pushes the boundaries of robot manipulation capability from several fronts:
1) it is an \textbf{ultra-long horizon} task where each step's success depends on the previous one;
2) the robot needs to perceive and manipulate \textbf{complex fluid} including both Newtonian fluid (i.e., water) and non-Newtonian fluid (i.e., ketchup).
3) the wiping motion requires using a \textbf{deformable tool} (i.e., sponge) while coordinating both arms with reference to the water stream;
4) manipulating \textbf{constrained articulated object} (i.e., turning on and off faucet) requires mechanical compliance provided by soft fingers;
5) the policy also need to be \textbf{semantically robust} to the concept of ``cleanliness''. When additional ketchup is added during washing or even after the washing phase is done, the robot needs to resume washing and wiping. 

\mypara{Performance} A single demonstrator collected 258 demonstrations with randomized initial states including, ketchup patterns, position of the plate and sponge, along with water faucet angle. The collected demo also include explicit demonstrations of recovery behavior when additional ketchup is added. For this task, we train diffusion policy by fine-tuning a CLIP \cite{radford2021learning} pretrained ViT-B/16 \cite{dosovitskiy2020image} vision encoder. 
Overall, UMI achieves $14/20=70\%$ success rate. In addition, we demonstrate the robustness of our policy against various distractors, and types of sauce (mustard, chocolate syrup, caramel syrup), as well as robustness against perturbations, see Fig. \ref{fig:UMI_robustness} and video on our \href{https://umi-gripper.github.io}{website} for details. 

\textbf{No CLIP-pretrained ViT vision encoder.}
For this visually complex task, we found training ResNet-34 \cite{he2015deep} from scratch to be insufficient. Specifically, the baseline policy with ResNet-34 learned an non-reactive behavior and ignored any variation in plate or sponge position. As a result, it cannot perform the task, $0/10=0\%$. 

\section{In-the-wild Generalization Experiments}
Prior works in behavior cloning typically only evaluate in the same environment as data collection, often limited by their inability to collect sufficiently diverse dataset to allow generalization. By not relying on teleoperation with real robots, UMI enables low-cost data collection in any environment, which we refer to as in-the-wild data. 

We evaluate UMI's ability to produce generalizable visuomotor policies by scaling up the cup arrangement task Sec. \ref{sec:eval_cup}  to \textbf{novel environments} and \textbf{novel objects}. 
Within 12 person-hours, 3 demonstrators collected 1400 demonstrations for the cup arrangement task across 30 diverse physical locations, including homes, offices, restaurants, and outdoor environments. The demonstrations involved 15 espresso cups of different colors, shapes (cylindrical and tapered), and materials (ceramic, glass, and metal). To ensure model capacity, we increased the vision encoder further to CLIP \cite{ravichandar2020recent} pretrained ViT-L/14 \cite{dosovitskiy2020image}.
We evaluated our UMI policy in two unseen environments (Fig. \ref{fig:in_the_wild_evaluation} first column):
\begin{itemize}
    \item \textbf{Cafe table} is a metal table in the outdoor seating area of a busy cafe where a large number of pedestrians serve as natural distractors. We tested 5 cups in the training set and 2 testing (unseen) cups, with 5 initial poses each, 35 experiments in total.
    \item \textbf{Water fountain} is a black cubic water fountain, with a thin film of water constantly flowing from the center, covering the entire top surface. This environment is notably out-of-distribution since all of our demonstrations are collected on non-black tables, not to mention changes in surface dynamics due to the presence of water. We tested 3 training and 2 testing cups, with 5 initial poses each, 25 experiments in total.
\end{itemize}
We selected the 2 testing cups such that one has an out-of-distribution color (dark blue), while the other has an unseen texture (brown rings). For each test case, we vary the initial pose of both the cup and the saucer. 

Our UMI policy has $28/40=70\%$ success rate on training cups and $15/20=75\%$ success rate on testing cups, with a combined success rate of $43/60=71.7\%$. More qualitative and quantitative results are shown in Fig. \ref{fig:in_the_wild_evaluation}.

\textbf{No in-the-wild data.}
To validate the generalization ability comes from in-the-wild data, instead of pretrained vision backbone, we trained another model that only uses data from our narrow-domain experiment collected in the same lab environment (described in Sec. \ref{sec:eval_cup}) with the same pretrained ViT vision backbone. 
In the same unseen environments, as shown in Fig. \ref{fig:in_the_wild_evaluation}(b), the robot with the baseline policy doesn't even move toward the cup. As a result, its success rate is 0\%. 

%
%

\textbf{Takeaway.} This result indicates that finetuning a large pre-trained model with narrow-domain data is insufficient for producing an in-the-wild deployable policy. Therefore collecting diverse, in-the-wild data is still critical for effective generalization to novel environments and objects.


\begin{figure*}[t]
\centering
\includegraphics[width=\linewidth]{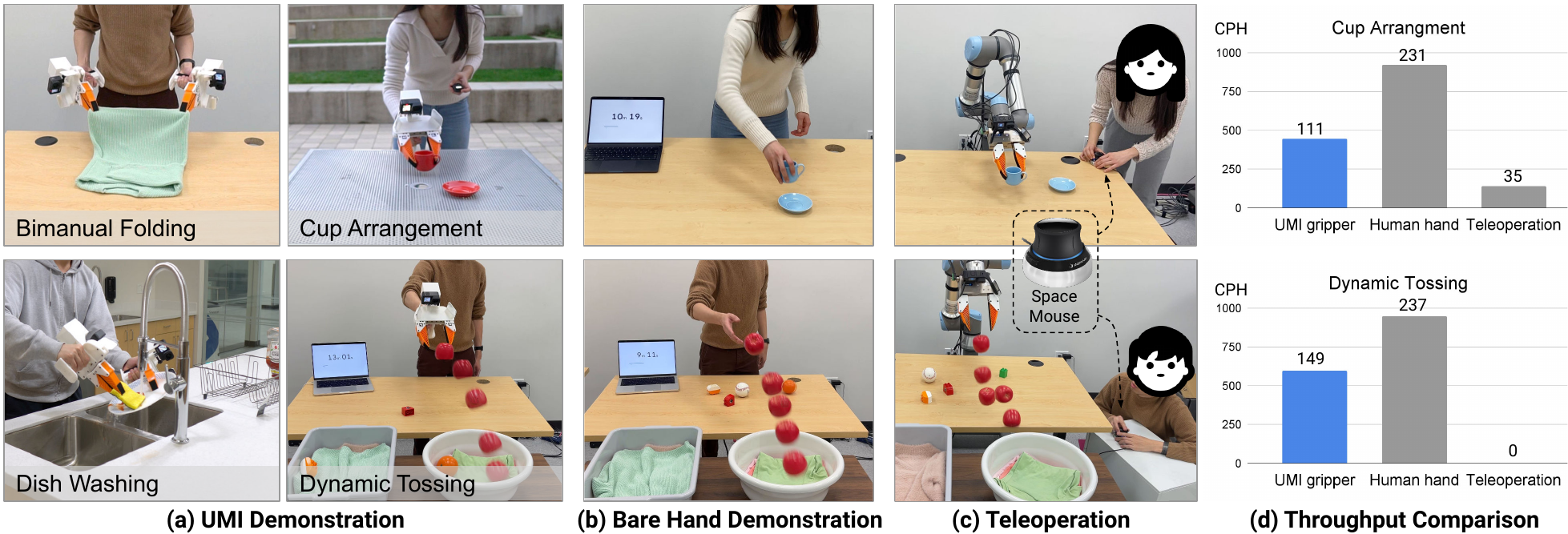}

\caption{\textbf{Data Collection Throughput Comparison} among UMI, bare hand, and teleoperation with a space mouse. UMI is not only faster than traditional teleoperation but also able to perform tasks that were not possible with typical teleoperation interfaces (e.g., dynamic tossing). 
}
\vspace{-3mm}
\label{fig:data_throughput}
\end{figure*}

\begin{figure}[t]
\centering
\includegraphics[width=\linewidth]{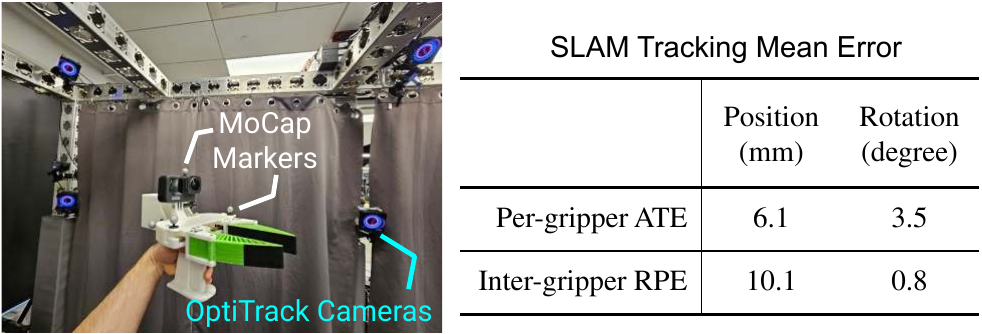}

\caption{\textbf{SLAM Accuracy.}  We evaluate the SLAM accuracy with a MoCap benchmark including 7 single-gripper tasks and 7 bimanual tasks. Overall, we can achieve less than 1 cm and 4$\degree$ tracking error. 
}
\vspace{-4mm}
\label{fig:slam}
\end{figure}

\section{Data Collection Throughput and Accuracy}
\label{sec:eval_thoughput}
\textbf{Throughput.} UMI's improved ergonomics and intuitiveness over teleoperations also lead to improved data collection throughput. 
To demonstrate this effect, we record the number of demonstrations that can be collected within 15 minutes by the same operation using 3 different methods (Fig. \ref{fig:data_throughput}): 
1) Human hand demonstration 
2) UMI gripper 
3) Spacemouse-based teleoperation, which is a typical teleoperation interface used in many learning from demonstration works \cite{chi2023diffusionpolicy, zhu2022viola}.  
We measure the data throughput on two tasks: 
1) cup arrangement
2) dynamic tossing.
Note that the time taken to reset the environment, randomize objects, and handle robot faults (such as self-collisions) are also counted in this experiment to accurately represent the real-world data collection throughput.

On the cup arrangement task, the UMI gripper is more than $3\times$ faster than teleportation, at $48\%$ speed of the human hand, shown in Fig. \ref{fig:data_throughput} (d). Note that human is significantly faster on reset and randomization, due to their proximity to the objects.
On the dynamic tossing task, the UMI gripper is at $64\%$ speed of the human hand, while the teleportation method failed to produce a single successful demonstration in 15 minutes.

\textbf{Accuracy.} To independently assess the accuracy of our SLAM-based tracking system, we collected a SLAM benchmark dataset with MoCap ground truth. The dataset contains 7 single-gripper tasks and 7 bimanual tasks, all with a variety of movable objects in view as well as natural and rapid human motion. 
As shown in Fig. \ref{fig:slam}, our SLAM system has a mean Absolute Trajectory Error (ATE) of $6.1mm$ for position and $3.5\degree$ for rotation. Since both grippers are localized with the same map, we can also obtain the relative pose between two grippers (i.e. inter-gripper pose \hyperref[PD2.3]{PD2.3}). The mean Relative Pose Error (RPE) between two grippers is $10.1mm$ for position and $0.8\degree$ for rotation.

\section{Limitations and Future Works}
While UMI demonstrates policy efficacy across a wide range of tasks and scenarios, a few limitations remain.
First, since the kinematics limits of the downstream deployment robots are unknown at the time of data collection, we rely on data filtering to ensure the kinematic feasibility of the resulting policy.
Future works could develop an embodiment-aware policy learning framework that can transfer skills from valid but hardware-infeasible actions.

Second, our SLAM-based action recovery system inherits visual SLAM's requirement for sufficient texture in the environment.
Future works could leverage static third-person-view cameras, coupled with additional fiducial markers on UMI grippers to recover action even in texture-deficient environments like rooms with pure white walls.

Third, collecting data with UMI grippers is still less efficient than human hand demonstration, as shown in Sec. \ref{sec:eval_thoughput}. This is in part due to the gripper's weight and bulkiness, and in part due to the reduced degrees of freedom compared to human hands.
Future works could explore lighter materials and further improve UMI gripper's mechanical design and ergonomics, or alternatively, build sufficiently capable dexterous robotic hands and policies that can directly transfer from human motions.



\section{Conclusion}
We present Universal Manipulation Interface (UMI), a framework that enables learning capable and generalizable manipulation policies directly from in-the-wild human demonstrations. 
The UMI gripper, a hand-held demonstration interface,
captures sufficient information to learn some challenging manipulation tasks, including washing a dirty dish, bimanual sweater folding, and dynamic object tossing and sorting.
At the same time, UMI remains highly scalable for in-the-wild data collection with its portability, cost-effectiveness, and operational simplicity.
By recording all information in a single, standardized MP4 file, UMI's data can be easily shared over the Internet, allowing geographically distributed data collection from a large pool of nonexpert demonstrators.
Our goal with UMI is to democratize robotic data collection, fostering a vast, diverse, and decentralized dataset to emerge from the robotics community

\section*{Acknowledgments}
This work was supported in part by the Toyota Research Institute, NSF Award \#2037101, and \#2132519. We want to thank Google and TRI for the UR5 robots, and IRIS and IPRL lab for the Franka robot hardware. The views and conclusions contained herein are those of the authors and should not be interpreted as necessarily representing the official policies, either expressed or implied, of the sponsors. 

We would like to thank Andy Zeng, Pete Florance, Huy Ha, Yihuai Gao, Samir Gadre, Mandi Zhao, Mengda Xu, Alper Canberk, Kevin Zakka, Zeyi Liu, Dominik Bauer, Tony Zhao, Zipeng Fu and Lucy Shi for their thoughtful discussions. We thank Alex Alspach, Brandan Hathaway, Aimee Goncalves, Phoebe Horgan, and Jarod Wilson for their help on hardware design and prototyping. We thank Naveen Kuppuswamy, Dale McConachie, and Calder Phillips-Graffine for their help on low-level controllers. We thank John Lenard, Frank Michel, Charles Richter, and Xiang Li for their advice on SLAM. We thank Eric Dusel, Nwabisi C., and Letica Priebe Rocha for their help on the MoCap dataset collection. We thank Chen Wang, Zhou Xian, Moo Jin Kim, and Marion Lepert for their assistance with the Franka setup. We especially thank Steffen Urban for his open-source projects on GoPro SLAM and Camera-IMU calibration, and John @ 3D printing world for inspiration of the gripper mechanism.


\bibliographystyle{plainnat}
\bibliography{references}

\newpage
\appendix
\renewcommand{\thesection}{A.\arabic{section}}
\renewcommand{\thefigure}{A\arabic{figure}}
\renewcommand{\thetable}{A\arabic{table}}
\setcounter{section}{0}
\setcounter{figure}{0}
\setcounter{table}{0}

Please check out our website (\href{https://umi-gripper.github.io}{https://umi-gripper.github.io}) for additional results and comparisons. In appendix, we present additional details on latency measurement \S \ref{sec:supp_latency_measurement}, data collection protocol \S \ref{sec:supp_data_collection}, evaluation protocol \S \ref{sec:supp_evaluation}, SLAM \S \ref{sec:supp_slam}, policy implementation \S \ref{sec:supp_policy}, and hardware implementation \S \ref{sec:supp_hardware}.

\subsection{\textbf{Latency Measurement}}
\label{sec:supp_latency_measurement}
As depicted in the main manuscript, accurate calibration of various latencies in the robotic system is critical for deploying UMI policies, especially for tasks that require rapid and dynamic actions. In the following sections, we will describe how we measure the latency for each component.

\subsubsection{\textbf{Camera Latency Measurement}}
\label{sec:camera_latency}
\begin{figure}[t]
    \centering
    \includegraphics[width=0.98\linewidth]{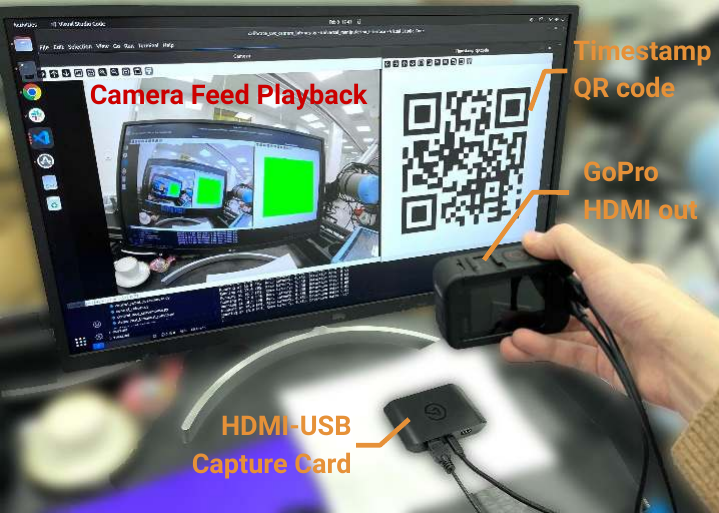}
    \caption{\textbf{Camera Latency Measurement}}
    \vspace{-4mm}
    \label{fig:camera_latency}
\end{figure}

For policy observation across both the UR5 and Franka FR2 platforms, we employ each robot arm with a single wrist-mounted GoPro Hero 9 camera. To obtain real-time video streams from the GoPro, we use a combination of GoPro Media Mod 1.0 (to convert usb-c to HDMI) and Elgato HD60X external capture card (to convert HDMI to USB-3.0 UVC interface). To measure the end-to-end latency of the camera pipeline, we record (with the GoPro camera) a rolling QR code on the computer monitor that displays the current system timestamp for each video frame $t_{display}$, as shown in Fig. \ref{fig:camera_latency}. To prevent multiple detections of QR codes through overlay of camera streams, we masked out the QR code in camera playback, as shown on the left of the monitor. By subtracting the receiving timestamp for each frame $t_{recv}$ and the decoded QR code timestamp $t_{display}$, and subtracting the known latency of display refresh $l_{display}$, we can obtain the end-to-end latency of camera system:
\[l_{camera}=t_{recv}-t_{display}-l_{display}\]

\subsubsection{\textbf{Proprioception Latency Measurement}}
\label{sec:proprioception_latency}
When the robotic hardware directly reports global timestamps, such is the case for Franka FR2 robot, we measure the proprioception latency by subtracting the robot sending timestamp $t_{robot}$ from the policy-received timestamp $t_{recv}$:
\[l_{obs}=t_{recv}-t_{robot}\]
When the robotic hardware timestamp is unavailable, such as the UR5 robot and Schunk WSG-50 gripper, we approximate the proprioception latency with $\frac{1}{2}$ of ICMP round-trip time (i.e. ping).

\subsubsection{\textbf{Gripper Execution Latency Measurement}}
\label{sec:gripper_latency}
To obtain the gripper execution latency $l_{action}$, we subtract the end-to-end latency $l_{e2e}$ by the proprioception latency $l_{obs}$. To measure $l_{e2e}$, we send a sequence of sinusoidal position commands to the gripper, and then record a sequence of gripper width preconceptions. The $l_{e2e}$ can be obtained by computing the optimal alignment between the desired gripper width signal and the signal of actual received gripper widths through cross-convolution.
\[l_{action}=l_{e2e}-l_{obs}\]

\subsubsection{\textbf{Robot Execution Latency Measurement}}
\label{sec:execution_latency}
Similar to the gripper, we also measure the execution latency of the robot (ether UR5 or Franka) by calculating $l_{e2e}$, as the optimal alignment between a sequence of desired end-effector poses and the measured actual end-effector poses. Due to safety concerns, we directly teleoperate the robot to generate the desired end-effector pose sequences.




\begin{table*}
\centering
\begin{tabular}{l|lllllllllllll}
\toprule
\textbf{H-Param}  & \textbf{I-To} & \textbf{P-To} & \textbf{Ta} & \textbf{Freq} & \textbf{Speed} & \textbf{ImgRes} & \textbf{V-Arch} & \textbf{V-Pretrain} & \textbf{D-Lr} & \textbf{V-Lr} & \textbf{Epochs} & \textbf{Batch} & \textbf{Compute} \\
\midrule
Cup Arrag.       & 2    & 2    & 6  & 10   & 0.5x  & 1x224x224 & ViT-B/16  & CLIP       & 3e-4 & 3e-5 & 250    & 512   & 4xA10g  \\
Obj Tossing      & 2    & 2    & 6  & 20   & 1.0x  & 1x224x224 & ResNet-34 & No         & 3e-4 & 3e-4 & 350    & 1024  & 4xA10g  \\
Cloth Folding    & 2    & 2    & 6  & 10   & 0.5x  & 2x224x224 & ResNet-34 & No         & 3e-4 & 3e-4 & 100    & 1024  & 4xA10g  \\
Dish Washing     & 2    & 2    & 6  & 10   & 0.5x  & 2x224x224 & ViT-B/16  & CLIP       & 3e-4 & 3e-5 & 90     & 224   & 8xA10g  \\
Cup (In-the-wild) & 1    & 2    & 6  & 10   & 0.5x  & 1x224x224 & ViT-L/14  & CLIP       & 3e-4 & 3e-5 & 50     & 512   & 8xA100 \\
\bottomrule
\end{tabular}
\caption{
\textbf{Hyperparameters for Diffusion Policy}
\label{tab:hparam}
\textbf{I-To}: image observation horizon.
\textbf{P-To}: proprioception observatino horizon.
\textbf{Ta}: action horizon.
\textbf{Freq}: environment frequency (Hz, both observation and action).
\textbf{Speed}: policy execution speed wrt data collection speed.
\textbf{ImgRes}: environment observation resolution (Camera views x W x H).
\textbf{V-Arch}: vision encoder architecture.
\textbf{V-Pretrain}: pretraining method for vision encoder.
\textbf{D-Lr}: learning rate for action diffusion model.
\textbf{V-Lr}: learning rate for vision encoder.
\textbf{Epochs}: training epochs.
\textbf{Batch}: total batch size over all GPUs.
\textbf{Compute}: number and type of GPUs used for training.
All tasks uses DDIM scheduler with 50 training diffusion steps and 16 inference steps.
}
\vspace{-3mm}
\end{table*}

\subsection{\textbf{Data Collection Protocol}}
\label{sec:supp_data_collection}
UMI enables in-the-wild data collection with close-to-zero setup time. To start data collection in a new environment, the demonstrator follows a 4-step process:
\begin{itemize}
    \item \textbf{Time Synchronization} (optional). When utilizing UMI in a bimanual confirmation, we synchronize the internal clocks for the two GoPro cameras by scanning a rolling QR code on a smartphone. ``GoPro Labs" experimental firmware recognizes the global timestamp encoded in the QR code and calibrates its internal clock with $\pm \frac{1}{60}$ second accuracy. This step can be skipped for single-arm tasks.

    \item \textbf{Gripper Calibration} (optional). We calibrate the minimum and maximum width between gripper fingers by recording a video of opening and closing the gripper for 5 times. While this step is only necessary once per gripper to account for manufacturing inconsistencies, we often do this for every data collection scene to avoid potential issues from accidentally mixing up between grippers between collection sessions.

    \item \textbf{Mapping}. For each new scene, we scan the environment by slowly moving the gripper around, following a pattern that first covers a sufficient part of the scene, then mimicking the likely motions for the desired task for denser coverage on task-relevant parts of the scene. As described in Sec. \ref{sec:supp_slam}, obtaining a high-quality map is critical for precise and robust SLAM tracking, especially for rapid motions during demonstrations. To further improve mapping robustness, we optionally place a fiducial marker on the table/ground which will be taken away during demonstration. The mapping process usually takes around 1 minute.

    \item \textbf{Demonstration}. In this step, the demonstrator performs the desired task repeatedly, within the same scene. We record one video per demonstration, with the video itself marks the start and end of each episode. Under single gripper configuration,  the demonstrator can press the recording button with the other hand on the GoPro, or optionally a GoPro remote control, to start and end each video recording. Under bimanual configuration,  the demonstrator uses GoPro's built-in voice command (\textit{``GoPro capture"}, \textit{``GoPro stop capture"}) to start and end video recording.
\end{itemize}

At the end of each shift/day, the demonstrator uploads the mp4 files on all grippers to cloud storage with one folder per scene. From there, we provide a single script that converts these mp4 files into a dataset that can be directly used to train diffusion policy. Under the hood, the script automatically detects and disambiguates left/right grippers using GoPros' serial numbers, matches the videos into synchronized pairs and recovers the robot actions using our SLAM pipeline and fiducial marker detection.

\subsection{\textbf{Evaluation Protocol}}
\label{sec:supp_evaluation}
In this section, we explain the process of real world policy evaluation in detail. Specifically, we will describe how we select \textbf{initial states}, how we \textbf{terminate} an experiment, and how we judge \textbf{success and failures}.

\subsubsection{\textbf{Initial State Selection}}
For all tasks, we manually select a set of initial states with diverse pose coverage across task scenes (for both the robot and the environment) that are shared across all evaluated methods. During evaluation, we manually match the initial states with a third-person camera to be close to pixel-perfect. We ensure the initial states to be kinematically feasible by teleoperating the robot. For narrow-domain tasks, we randomize the objects with roughly the same distribution as data collection. 

\subsubsection{\textbf{Termination Criteria}}
During evaluation, an operator supervises the robot at all times. An evaluation episode can be terminated due to:
\begin{itemize}
    \item \textbf{Safety Concern.} When the operator deems the robot is about to perform dangerous actions that could potentially break the setup/robot or do any other harm, the episode will be terminated immediately.

    \item \textbf{Robot Fault.} When the robot enters a fault state, potentially due to external or self-collisions, excessive force, or joint/speed limit violation, the episode is terminated automatically.

    \item \textbf{Timeout.} We manually set a task-specific conservative timeout, usually 3-5 times longer than the median task duration, to automatically terminate the episode. This timeout is usually triggered when the policy is stuck and stops making progress.

    \item \textbf{Task Success.} The operator manually terminates the episode when the task is deemed to be successful. 
\end{itemize}

Since the termination criteria contain subjective elements, we encourage readers to check out our complimentary video that contains all experiments reported.

\subsubsection{\textbf{Success Criteria}}
It is difficult to define automatic and compact success metrics for complex manipulation tasks reported in this paper. Therefore, the operator manually judges the success or failure of each episode using the rubric described below. While we try to create a concise and objective rubric, it inevitability contains subjective elements. As a result, viewing our complimentary video submission is still the best way to elucidate the relative performance between our method and various baselines.

\textbf{Cup Arrangement.} We define task success as when the cup is placed upright on the saucer with its handle within ±15° to the left of the robot.

\textbf{Dynamic Tossing.} We define task success as when the objects are tossed into their corresponding bins. Bouncing off from inside the bin is counted as success, as long as the object hits the bottom of the bin. If an object hits the sidewall of the bin and bounces back to the table, then successfully tossed in the bin later, it also counts as a success.

\textbf{Bimanual Cloth Folding.} We define task success as when both sleeves are folded inside of the sweater, and the silhouette (projected area) of the folded sweater resembles a square, sized around a quarter of area of the sweater torso. A sleeve folded in but not perfectly flat still counts as success. Even if the cloth is not perfectly flat, the episode is still deemed successful as long as the projected area resembles a square.

\textbf{Dish Washing.} An episode is deemed successful when the plate is clean and placed vertically back on the rack. Residual spots of ketchup with a size <5mm are deemed clean. If the sponge is placed on the edge of the sink and then slips inside the sink, it does not affect task success as long as the plate is clean and placed back on the rack.

\begin{figure}[t]
    \centering
    \includegraphics[width=0.98\linewidth]{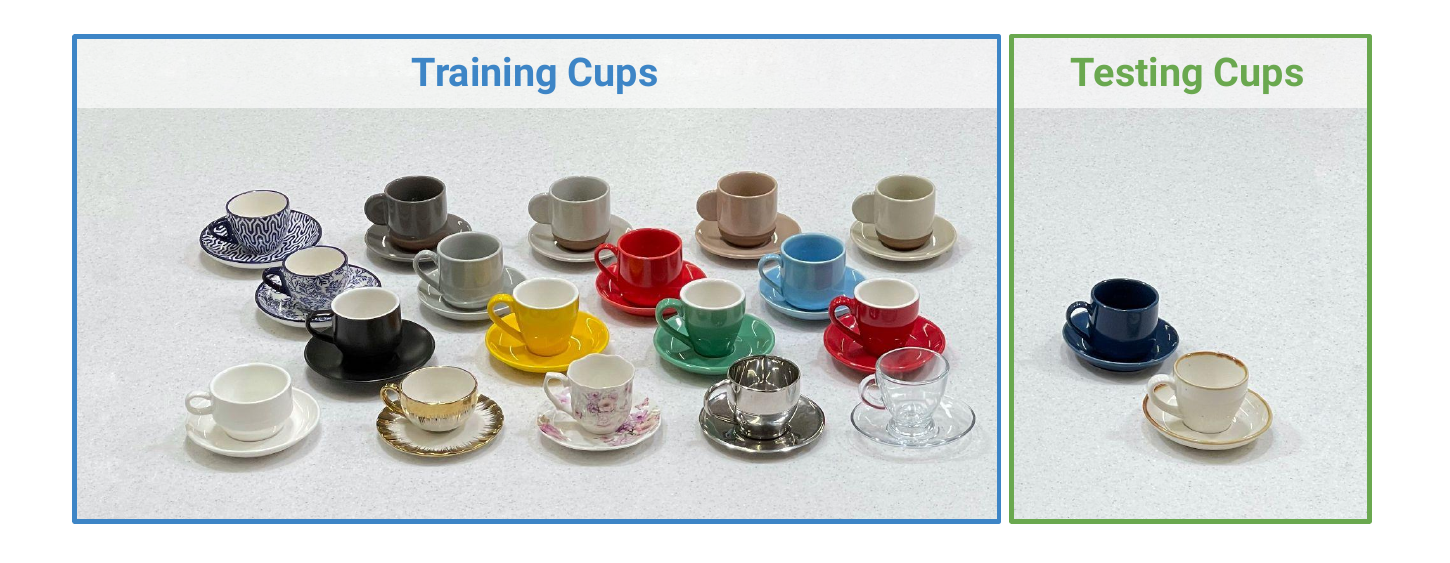}
    \caption{\textbf{Cup Selections.} For the in-the-wild cup arrangement task, we purchased 20 different styles of ``espresso cup with saucer" on Amazon, of which 18 are used for training and 2 are reserved for testing.} 
    \label{fig:cup-distribution}
    \vspace{-3mm}
\end{figure}

\textbf{In-the-wild Cup Arrangement.} The success criterion is the same as the narrow-domain cup arrangement, except the concept of ``left" is defined with respect to the initial pose of the gripper for each episode. See Fig. \ref{fig:cup-distribution} for our selection of training and testing cups.

\subsection{SLAM System Details}
\label{sec:supp_slam}
We found OBR-SLAM3's \cite{orbslam3} original monocular-inertial SLAM system to be inadequate for our application. In particular, ORB-SLAM3 has an initialization process where the first few map points and key frames, as well as IMU calibration parameters, are heuristically computed. We found this initialization process to be particularly fragile and lengthy under fast movements, during which no camera pose can be estimated, resulting in a large proportion of wasted data. To solve this problem, we implemented two features: \textbf{Map as Initialization} and \textbf{Marker-enhanced Initialization}.

\textbf{Map as Initialization.} The original ORB-SLAM3 has a localization mode that localizes to an existing map without changing the map in any way, including creating more map points or performing any global optimization. We found the existing localization mode to be insufficiently robust since UMI manipulation dynamically changes the scene. To solve this problem, we modified ORB-SLAM3 to continue normal SLAM operation after relocalizing to an existing map loaded from disk, effectively using the existing map as only initialization for optimization. 

\textbf{Marker-enhanced initialization.} Due to inherent ambiguities of the monocular SLAM formulation, ORB-SLAM3's existing initialization struggles when features are far away (outdoor environment) or with large numbers of feature mismatches (repeated patterns, trees, etc.). We modified ORB-SLAM3 to optionally take advantage of fiducial markers \cite{aruco} with known sizes to disambiguate possible explanations of feature matches. We found this feature to significantly increase mapping robustness in-the-wild. Note that demonstration videos will not contain these fiducial markers, they are only used for mapping.

\subsection{\textbf{Policy Implementation Details}}
\label{sec:supp_policy}
We use Diffusion Policy~\cite{chi2023diffusionpolicy} for all tasks. Detailed hyperparameters are listed in Tab. \ref{tab:hparam}. Compared to the original Diffusion Policy, here are some key modifications:

\subsubsection{\textbf{Vision encoder}} We utilize the Vision Transformer (ViT)~\cite{dosovitskiy2020image} as the vision encoder due to its substantial capacity in comparison to ResNet~\cite{he2015deep}, which proves crucial for tasks demanding intricate perceptual capabilities. Notably, the dataset collected for each task lacks the scale required for training ViT from scratch. To address this limitation, we employ the pre-trained CLIP ViT and implement a reduced learning rate, set at 10 times smaller during the fine-tuning process. While ViT-B suffices for most in-domain tasks, ViT-L can further enhance the performance of the in-the-wild cup arrangement task, which involves diverse environments and objects. To expedite the training and inference of ViT-L, we exclusively use a one-step observation as input, deviating from the two-step observation window approach employed in other tasks.

\subsubsection{\textbf{Frequency}} For most quasi-static tasks, a frequency of 10Hz proves sufficient for both observation and action. However, a frequency of 20Hz is employed for the dynamic tossing task, which requires highly reactive behaviors.

\subsubsection{\textbf{Speed}} The output of Diffusion Policy is a sequence of actions, specifically the target pose, with an implicit $dt_{output}$ between two steps determined by the demonstration dataset. However, during execution, we are not bound to follow the same $dt$. By adjusting the $dt_{execution}$, we can achieve different execution speeds compared to the human demonstration. In the case of the dynamic tossing task, maintaining the original speed is essential to ensure the object attains sufficient releasing velocity. However, for other quasi-static tasks, we observed that a 0.5x slower speed results in smoother behavior. This phenomenon may arise from imperfect latency compensation and can potentially be mitigated through improved latency matching.

\subsubsection{\textbf{Image Augmentation}} We employ a set of image augmentations to enhance the diversity of our training data, thereby improving the robustness and generalization capabilities of our policy. The augmentation pipeline includes a RandomCrop operation with a ratio of 0.95, a RandomRotation operation with degrees ranging from -5.0 to 5.0, and a ColorJitter operation, adjusting brightness by up to 0.3, contrast by 0.4, saturation by 0.5, and hue by 0.08. The specific parameter choices are informed by the distribution of our dataset.

\subsection{Hardware Implementaiton Details}
\label{sec:supp_hardware}

\subsubsection{Soft Compliant Fingers} 
\begin{figure}[t]
    \centering
    \includegraphics[width=0.6\linewidth]{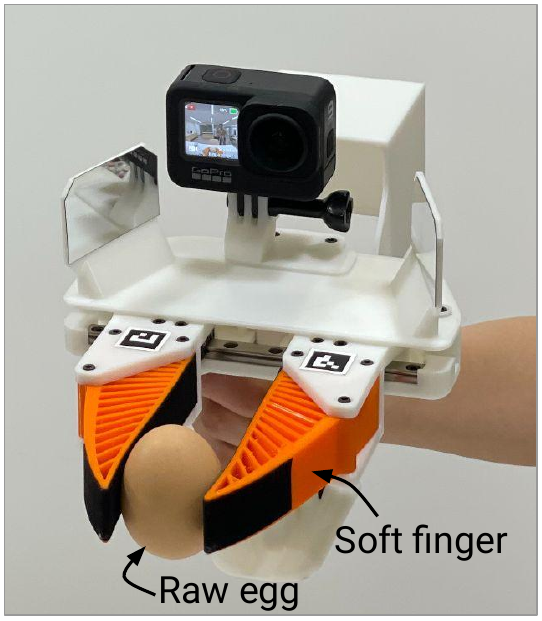}
    \caption{\textbf{Soft Fingers} printed with TPU material provides passive mechanical compliance, enabling a secure grip on a raw egg without causing breakage.}
    \vspace{-3mm}
    \label{fig:soft-finger}
\end{figure}

We used the same soft fingers on both UMI data collection grippers as well as deployed robotic grippers. Printed with 95A TPU material, the rib-like pattern on the finger maintains rigidity on the fingertip while conforming to the object geometry for a more secure grasp (Fig. \ref{fig:soft-finger}). When deployed to robots that lack force-torque control such as UR5, the deformable nature of our soft fingers provides passive mechanical compliance, critical for contact-rich tasks such as opening a water faucet and scooping up thin clothes from a table.
The soft fingers also provide some implicit grasping force control when combined with continuous gripper control.

\subsubsection{Franka Mount}
Due to FR2's limited end-effector pitch (FR2 is designed for top-down pick and place, while the UMI gripper is mostly held horizontally), we had to design and 3D print a custom mounting adapter that rotates WSG50 gripper 90-degree rotation with respect to the robot's end-effector flange.

\end{document}